\documentclass{article}
\usepackage[preprint]{neurips_2025}
\usepackage[utf8]{inputenc}
\usepackage{placeins}
\usepackage[T1]{fontenc}

\usepackage{enumitem}
\usepackage{url}
\usepackage{booktabs}
\usepackage{amsfonts}
\usepackage{amsmath} 
\usepackage{amssymb}  
\usepackage{nicefrac}
\usepackage{xcolor}
\usepackage{graphicx}
\usepackage{wrapfig}  
\usepackage{multirow}
\usepackage[hidelinks]{hyperref}

\usepackage{needspace}

\usepackage{float}
 
\usepackage{amsmath,amssymb}

\title{Semantic Anchors in In-Context Learning: Why Small LLMs Cannot Flip Their Labels}

\author{
Anantha Padmanaban Krishna Kumar \\
Department of Computer Science, Boston University\\
\texttt{anantha@bu.edu} \\
}

\begin{document}
\maketitle

\begin{abstract}
Can in-context learning (ICL) override pre-trained label semantics, or does it merely refine an existing semantic backbone? We address this question by treating LLMs as prompt-induced classifiers and contrasting their behavior under \emph{natural} demonstrations (with correct labels) and \emph{inverted} demonstrations (systematically flipping label meanings). We decompose ICL behavior into three alignment metrics (truth, prior, and prompt alignment) and introduce a semantic override rate, defined as correctness under flipped semantics. Across eight classification tasks and eight open-source LLMs (1--12B parameters), we find consistent evidence for a semantic anchor view. With natural demonstrations, ICL improves accuracy while maintaining strong prior alignment; most correct predictions coincide with zero-shot behavior, even when the prior is weak. With inverted demonstrations, models cannot learn coherent anti-semantic classifiers: prompt alignment increases only by sacrificing accuracy, and semantic override rates remain exactly zero in our few-shot 1--12B setting. Rather than flexibly remapping label meanings, ICL primarily adjusts how inputs project onto stable semantic directions learned during pre-training, clarifying fundamental limits of few-shot prompting and suggesting that overriding label semantics at these scales requires interventions beyond ICL. All code is available at: \url{https://github.com/AnanthaPadmanaban-KrishnaKumar/semantic-anchors-icl}.
\end{abstract}

\section{Introduction}
Large language models (LLMs) exhibit remarkable in-context learning (ICL): given natural-language instruction and a few input-label examples, frozen models generalize to new inputs without parameter updates \citep{brown2020language}. This gradient-free adaptation has made ICL the standard paradigm for LLM deployment, sometimes approaching fine-tuned performance at zero training cost. Yet, the mechanism remains contested: does ICL learn new input-label mappings, or merely refine pre-trained behavior?

Two theories compete. The \emph{task learning} view treats ICL as a general learning algorithm that is either implicit Bayesian inference \citep{xie2021explanation} or gradient descent simulation \citep{dai2023can,akyurek2022learning,von2023transformers}, and that flexibly adopts any coherent mapping from consistent demonstrations. The \emph{prior refinement} view counters that ICL sharpens existing classifiers rather than learning de novo: random label permutations barely hurt performance \citep{min2022rethinking}, ICL violates Bayesian consistency \citep{falck2024context,kossen2023context}, and methods like Self-ICL successfully bootstrap from zero-shot predictions alone.

At stake is whether pre-trained \emph{semantic anchors} can be overridden. Label tokens carry deep semantic commitments: changing \texttt{positive} to \texttt{great} can swing accuracy by tens of points \citep{schick2021exploiting,gao-etal-2021-making}, and models exhibit systematic biases toward common tokens \citep{zhao2021calibrate,fei-etal-2023-mitigating}. \citet{wei2023larger} showed that overriding these anchors (making models label positive reviews as \texttt{NEG}) requires massive scale: GPT-3 can eventually flip labels, but smaller models cannot. Whether the 1--12B models that dominate open deployments can learn anti-semantic mappings remains unclear for today's open-source families and across tasks in the few-shot regime. Unlike prior flipped-label studies focused on large proprietary models or narrow task sets, we systematically map this effect across small open-source families and quantify it with alignment metrics.

We test semantic flexibility directly. Using \emph{natural} demonstrations (with correct labels) versus \emph{inverted} demonstrations (systematically flipped labels), we decompose ICL into three alignments: truth (accuracy), prior (zero-shot agreement), and prompt (agreement with the demonstrated mapping). Our key metric, the \emph{semantic override rate}, counts predictions that are correct under the inverted mapping. Across eight tasks and eight open-source LLMs (1--12B parameters across the LLaMA, Mistral, Qwen, and Gemma families), we find two consistent patterns. \textbf{Natural ICL refines priors:} accuracy improves while maintaining tight zero-shot coupling, even when priors are weak. \textbf{Inverted ICL fails to remap semantics:} models partially follow inverted demonstrations but never learn the intended anti-semantic classifier, and the semantic override rate is zero, not near-zero but exactly zero across thousands of predictions.

ICL adjusts how inputs project onto a pre-trained semantic space, but cannot redefine what labels mean. These rigid semantic anchors explain both ICL's effectiveness (when aligned with pre-training) and its fundamental limits (when opposed to it).


\section{Related Work}
\label{sec:related}

\textbf{Semantic manipulation in ICL.} Prior work has tested ICL's flexibility through label manipulation. \citet{min-etal-2022-rethinking} found that random label permutations cause only marginal accuracy drops, though this tests noise robustness rather than coherent remapping. \citet{wei2023larger} directly tested semantic override with flipped labels (positive reviews labeled \texttt{NEG}), finding a stark scale dependency: GPT-3-sized models can eventually adopt inverted mappings, but smaller models cannot. \citet{agarwal2024many} showed that even many-shot regimes (hundreds of examples) struggle with semantic override at smaller scales. Related analyses attribute flipped-label failure to demonstration shortcuts or prior dominance in ICL. We extend this inquiry to the 1--12B parameter range with explicit alignment metrics that decompose the failure modes rather than reporting accuracy alone.

\textbf{Theoretical frameworks.} Two mechanistic theories dominate. Bayesian accounts frame ICL as inference over latent tasks \citep{xie2021explanation,panwar2023context}, while meta-optimization views argue that transformers simulate gradient descent \citep{akyurek2022learning,von2023transformers,dai2023can}. Both often model labels as arbitrary symbols, implying that they should be remappable under consistent demonstrations. However, \citet{falck2024context} showed that ICL violates Bayesian consistency, and \citet{kossen2023context} demonstrated that ``label relationships inferred from pre-training have a lasting effect that cannot be surmounted by in-context observations.'' These findings support a prior-constrained learning view rather than flexible remapping.

\textbf{Label semantics as constraints.} Labels carry strong semantic priors: switching \texttt{positive} to \texttt{great} can change accuracy by tens of points \citep{schick2021exploiting,gao-etal-2021-making,cui-etal-2022-prototypical,mueller-etal-2022-label}. \citet{zhao2021calibrate} identified systematic biases (majority-label, recency, common-token) that require explicit calibration. \citet{holtzman-etal-2021-surface} showed that semantically equivalent forms compete for probability mass. This evidence frames labels as semantic anchors rather than neutral symbols, a view implicitly acknowledged by methods like Self-ICL that bootstrap from zero-shot predictions. When override is necessary, approaches such as symbol tuning \citep{wei2023symbol} or contrastive decoding \citep{peng2025enhancing} bypass these constraints through fine-tuning or inference-time interventions, suggesting that standard ICL alone cannot overcome semantic anchors.


\section{Problem Setup}
\label{sec:problem-setup}

We formalize in-context learning as a choice between two classifiers: a zero-shot prior and an in-context classifier induced by demonstrations.

\subsection{Zero-shot and In-Context Classifiers}
\label{subsec:classifiers}

Let $\mathcal{X}$ denote the input space and $\mathcal{Y}$ the label set. Each input $x \in \mathcal{X}$ has a ground-truth label $y^*(x) \in \mathcal{Y}$. Given a causal language model and a prompt $P$, we obtain predictions by greedy decoding.

The \textbf{zero-shot prior} $f_0(x) \in \mathcal{Y} \cup \{\textsc{unk}\}$ uses only task instructions, capturing pre-trained tendencies. The \textbf{in-context classifier} $f_{\text{icl}}(x; S) \in \mathcal{Y} \cup \{\textsc{unk}\}$ conditions on a demonstration set $S = \{(x_i, y_i)\}_{i=1}^k$. We evaluate all metrics on $\mathcal{D}_k = \{x \mid f_{\text{icl}}(x; S_k) \neq \textsc{unk}\}$.

\subsection{Natural vs. Inverted Demonstrations}
\label{subsec:nat-inv}

We contrast two demonstration regimes. \textbf{Natural} demonstrations use correct labels: $S_{\text{nat}} = \{(x_i, y^*(x_i))\}_{i=1}^k$. \textbf{Inverted} demonstrations apply a permutation $\phi : \mathcal{Y} \to \mathcal{Y}$ to labels: $S_{\text{inv}} = \{(x_i, \phi(y^*(x_i)))\}_{i=1}^k$. For sentiment, $\phi$ swaps POS $\leftrightarrow$ NEG; for NLI, $\phi$ cycles ENTAILMENT $\rightarrow$ NEUTRAL $\rightarrow$ CONTRADICTION $\rightarrow$ ENTAILMENT.

The prompt-favored label is $y_{\text{prompt}}(x) = y^*(x)$ under natural demonstrations and $\phi(y^*(x))$ under inverted demonstrations.

\subsection{Alignment Metrics}
\label{subsec:alignment}

To decompose how demonstrations affect predictions, we measure three alignments:
\begin{align}
    \text{Accuracy}(k) &= \frac{1}{|\mathcal{D}_k|} \sum_{x \in \mathcal{D}_k} \mathbf{1}[f_{\text{icl}}(x; S_k) = y^*(x)] \label{eq:acc}\\
    \text{Prior Alignment}(k) &= \frac{1}{|\mathcal{D}_k|} \sum_{x \in \mathcal{D}_k} \mathbf{1}[f_{\text{icl}}(x; S_k) = f_0(x)] \label{eq:prior}\\
    \text{Prompt Alignment}(k) &= \frac{1}{|\mathcal{D}_k|} \sum_{x \in \mathcal{D}_k} \mathbf{1}[f_{\text{icl}}(x; S_k) = y_{\text{prompt}}(x)] \label{eq:prompt}
\end{align}

\textbf{Prior alignment} reveals whether ICL refines or overrides zero-shot behavior. \textbf{Prompt alignment} measures agreement with demonstrated mappings. Under natural demonstrations, prompt alignment equals accuracy. Under inverted demonstrations, these metrics diverge.

Our key metric is the \textbf{semantic override rate}: the probability that predictions are both correct and consistent with inverted mappings, $P[f_{\text{icl}}(x; S_k) = y^*(x) \land f_{\text{icl}}(x; S_k) = y_{\text{prompt}}(x)]$ under inverted demonstrations. This captures true semantic flexibility: whether models can accurately apply anti-semantic rules.

\section{Experimental Setup}
\label{sec:experimental-setup}

\subsection{Models}
\label{subsec:models}
\begin{wraptable}{r}{0.55\textwidth}
\centering
\vspace{-0.5cm}
\small
\begin{tabular}{l|l}
\toprule
Model Family & Model Name (Abbr.) \\
\midrule
\multirow{3}{*}{LLaMA} & LLaMA-3.1-8B-Base (L8B) \\
                       & LLaMA-3.1-8B-Inst (L8I) \\
                       & LLaMA-3.2-3B-Inst (L3I) \\
\midrule
Mistral & Mistral-7B-Inst-v0.3 (M7I) \\
\midrule
Qwen & Qwen2.5-7B (Q7) \\
\midrule
\multirow{3}{*}{Gemma} & Gemma-3-1B-IT (G1I) \\
                       & Gemma-3-4B-IT (G4I) \\
                       & Gemma-3-12B-IT (G12I) \\
\bottomrule
\end{tabular}
\caption{Models evaluated}
\label{tab:models}
\vspace{-0.5cm}
\end{wraptable}

We evaluated eight models in four architectural families that span 1--12B parameters: LLaMA \citep{grattafiori2024llama}, Mistral \citep{jiang2023mistral7b}, Qwen \citep{yang2025qwen3}, and Gemma \citep{team2025gemma}. This diverse set eliminates architectural confounds: Gemma uses different positional encodings, Qwen employs distinct tokenization, and the 12× scale range tests whether semantic anchoring weakens with capacity. The base versus instruction-tuned comparison (LLaMA 3.1 8B) isolates the effect of fine-tuning on semantic rigidity. All models are accessed via Hugging Face with identical evaluation pipelines.

\subsection{Tasks and Datasets}
\label{subsec:datasets}

We evaluated eight classification tasks where the label tokens have strong semantic meaning: \textbf{Sentiment Analysis} (\citealt{socher-etal-2013-recursive}, SST-2; \citealt{maas-etal-2011-learning}, IMDB) with \{\texttt{POS}, \texttt{NEG}\}; \textbf{Natural Language Inference} (\citealt{bowman-etal-2015-large}, SNLI; \citealt{williams-etal-2018-broad}, MNLI) with \{\texttt{ENTAILMENT}, \texttt{NEUTRAL}, \texttt{CONTRADICTION}\}; \textbf{Paraphrase Detection} (\citealt{dolan-brockett-2005-automatically}, MRPC; \citealt{quora_question_pairs}, QQP); \textbf{Hate Speech Detection} (\citealt{mollas2022ethos}, ETHOS); and \textbf{Topic Classification} (\citealt{zhang2015character}, AG News).

\subsection{Prompting Conditions}
\label{subsec:prompting}

\begin{table}[ht]
\centering
\small
\setlength{\tabcolsep}{4pt}
\begin{tabular}{@{}p{0.30\textwidth} p{0.30\textwidth} p{0.30\textwidth}@{}}
\toprule
\textbf{Zero-shot} & \textbf{Natural $k$-shot} & \textbf{Inverted $k$-shot} \\
\midrule
\texttt{You are a sentiment classifier.} \newline
\texttt{Classify the sentiment of the following review as POS or NEG.} \newline\newline
\texttt{Review: <text>} \newline
\texttt{Label:} & 

\texttt{You are a sentiment classifier.} \newline
\texttt{Classify the sentiment of the following review as POS or NEG.} \newline\newline
\texttt{Review: <$x_1$>} \newline
\texttt{Label: POS} \newline
\texttt{Review: <$x_2$>} \newline
\texttt{Label: NEG} \newline
\texttt{...} \newline
\texttt{Review: <text>} \newline
\texttt{Label:} & 

\texttt{You are a sentiment classifier.} \newline
\texttt{Classify the sentiment of the following review as POS or NEG.} \newline\newline
\texttt{Review: <$x_1$>} \newline
\texttt{Label: \textcolor{red}{NEG}} \newline
\texttt{Review: <$x_2$>} \newline
\texttt{Label: \textcolor{red}{POS}} \newline
\texttt{...} \newline
\texttt{Review: <text>} \newline
\texttt{Label:} \\
\bottomrule
\end{tabular}
\caption{Example prompting conditions for sentiment classification. Natural $k$-shot uses correct labels; inverted $k$-shot flips labels, shown in red. Other tasks follow the same pattern.}
\label{tab:prompting}
\end{table}

We evaluate three prompting conditions, illustrated in Table~\ref{tab:prompting}. For each test example, we sample $k$ demonstrations from the same dataset using fixed random seeds to ensure consistent comparisons across models and conditions.

\subsection{Implementation Details}
\label{subsec:implementation}

We generate up to 3 tokens after the label stub using greedy decoding (temperature 0) and map the first generated word to labels using task-specific heuristics. Outputs that do not match any label are marked as \textsc{unk} and excluded from metrics. We test with $k \in \{1, 2, 4, 8\}$ demonstrations, using identical demonstration sets across all models. Input length is capped at 512 tokens for zero-shot and 2,048 for few-shot prompts. All experiments use 5 random seeds (0, 1, 2, 42, 123) for demonstration sampling. We evaluate the three alignment metrics plus the semantic override rate for binary tasks.


\section{Results}
\label{sec:results}

\begin{figure}[ht]
  \centering
  \includegraphics[width=\linewidth]{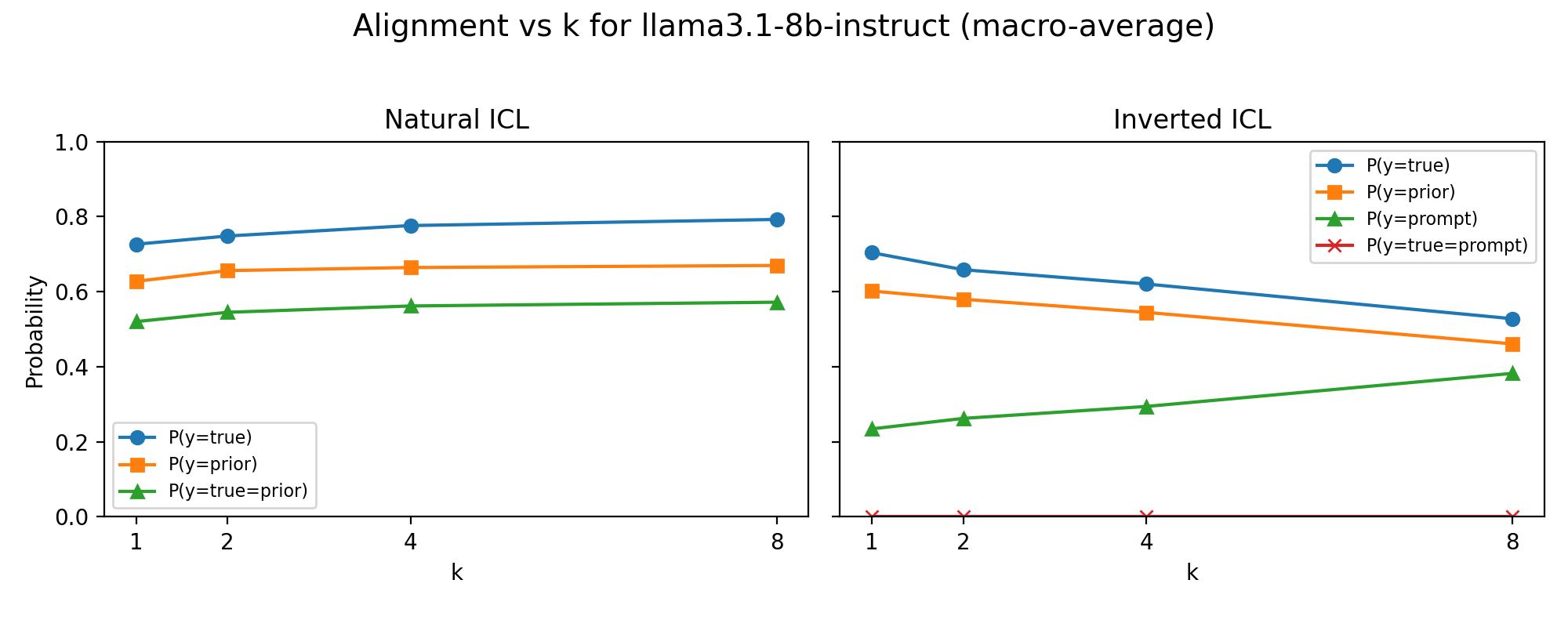}
  \caption{Macro-averaged alignment probabilities for LLaMA-3.1-8B-Instruct. Left: natural ICL increases both accuracy and joint correctness while maintaining prior alignment. Right: inverted ICL degrades accuracy and prior alignment as prompt-following increases, but joint alignment remains zero.}
  \label{fig:llama-align-vs-k}
\end{figure}

We analyze in-context learning behavior for LLaMA-3.1-8B-Instruct across eight classification tasks, contrasting natural demonstrations (correct labels) with inverted demonstrations (systematically flipped labels). Results for all eight models are provided in Appendix~\ref{sec:appendix_results}.

\subsection{Natural ICL improves accuracy, inverted ICL degrades it}

Table~\ref{tab:l8i-main} reveals a fundamental asymmetry. Natural ICL improves average accuracy from 69.5\% (zero-shot) to 79.3\% at $k=8$. Gains are largest where zero-shot priors are weakest: QQP jumps 37.8 points (40.6\% $\rightarrow$ 78.4\%), while NLI tasks gain 14--17 points. Even strong baselines improve: sentiment classification adds 2--3 points despite starting above 90\%.

Inverted ICL produces the opposite effect. Average accuracy drops to 52.8\% at $k=8$, with severe degradation on semantically rich tasks. Sentiment classification loses over 40 points; topic classification drops 32 points. The sole exception is QQP, where inverted ICL (71.6\%) exceeds the weak zero-shot baseline yet still underperforms natural ICL by 7 points. This suggests that when priors are sufficiently weak, even corrupted demonstrations provide task-relevant signal, though never enough to override semantic anchors.

\begin{figure}[ht]
  \centering
  \includegraphics[width=\linewidth]{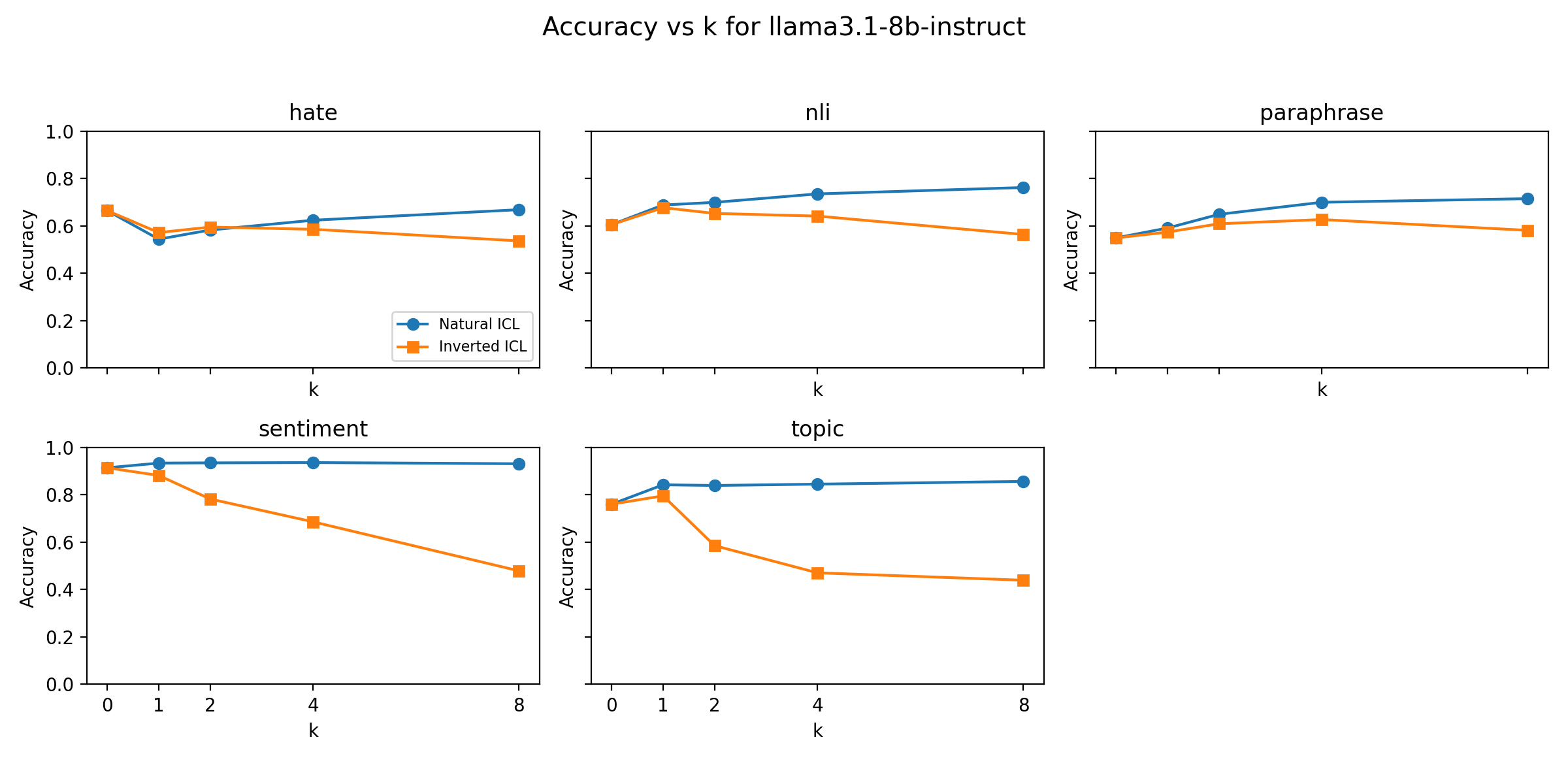}
  \caption{Accuracy vs.\ demonstrations $k$ for LLaMA-3.1-8B-Instruct. Natural ICL (blue) improves performance; inverted ICL (orange) degrades systematically with more examples.}
  \label{fig:llama-acc-vs-k}
\end{figure}

\subsection{ICL operates through prior refinement, not replacement}

To understand the mechanism, we decompose ICL behavior through alignment analysis (Table~\ref{tab:l8i-align}). Under natural demonstrations, $k$-shot predictions remain strongly coupled to zero-shot behavior. SST-2 achieves 92.5\% accuracy at $k=8$ while maintaining 89.8\% alignment with zero-shot predictions: 86.4\% of examples are both correct and consistent with the prior.

This coupling persists even when ICL dramatically improves accuracy. On QQP, where accuracy increases by 37 points, the model still agrees with its zero-shot prior on 34.9\% of examples. Figure~\ref{fig:llama-align-vs-k} (left) shows that as $k$ increases, both accuracy $P(y_{\text{icl}} = y^*)$ and joint correctness $P(y_{\text{icl}} = y^* = y_0)$ rise together, while raw prior alignment $P(y_{\text{icl}} = y_0)$ remains stable. Natural ICL sharpens correct predictions while suppressing errors: refinement rather than replacement.

\subsection{The semantic override paradox}

The critical test comes with inverted demonstrations. If ICL learned arbitrary input-output mappings, consistent anti-semantic examples should induce coherent label flipping. Instead, we observe a paradox: models partially follow inverted demonstrations but never learn the intended remapping.

Table~\ref{tab:l8i-align} quantifies this precisely. On sentiment tasks, roughly half of predictions match the demonstrated (incorrect) labels: $P(y = \tilde{y})$ reaches 51.6\% on IMDB and 52.6\% on SST-2. Yet the semantic override rate, $P(y = y^* = \tilde{y})$, remains exactly zero. No examples simultaneously satisfy both the true label and the inverted mapping.

Figure~\ref{fig:llama-align-vs-k} (right) reveals the mechanism: under inversion, accuracy and prior alignment decay while prompt-following $P(y = \tilde{y})$ increases, but joint alignment $P(y = y^* = \tilde{y})$ stays at zero. The model cannot reconcile conflicting constraints (demonstrated mappings versus semantic priors), producing incoherent outputs that satisfy neither.

\subsection{Effects strengthen with demonstration count}

The divergence between natural and inverted ICL intensifies with more demonstrations (Figure~\ref{fig:llama-acc-vs-k}). Natural ICL shows monotonic improvement or stability in all tasks as $k$ increases from 1 to 8. Gains correlate inversely with zero-shot strength: weak priors (QQP) show steep improvement, while strong priors (SST-2) show modest gains.

Inverted ICL exhibits the opposite pattern. Sentiment tasks that barely degrade at $k=1$ (losing 5--10 points) collapse at $k=8$ (losing more than 40 points). This is not random variation, but a systematic conflict: each additional inverted example strengthens the contradiction between demonstrated and pre-trained semantics, forcing predictions further from both the prior and ground truth.


\begin{table}[t]
  \centering
  \small
  \caption{
    LLaMA-3.1-8B-Instruct on eight datasets.
    ``Zero-shot'' reports accuracy and unknown rate without demonstrations.
    ``Natural'' and ``Inverted'' report $k{=}8$ ICL accuracy.
    All numbers are percentages and use only non-unknown predictions for accuracy.
  }
  \label{tab:l8i-main}
  \begin{tabular}{llrrrr}
    \toprule
    Task & Dataset &
    Zero-shot Acc. & Unknown &
    Natural Acc. ($k{=}8$) &
    Inverted Acc. ($k{=}8$) \\
    \midrule
    Hate        & ETHOS     & 66.4 &  0.2 & 66.8 & 53.7 \\
    NLI         & MNLI      & 60.9 &  0.0 & 77.6 & 58.9 \\
    NLI         & SNLI      & 60.2 &  0.0 & 74.8 & 53.9 \\
    Paraphrase  & MRPC      & 69.2 &  7.1 & 64.6 & 44.7 \\
    Paraphrase  & QQP       & 40.6 &  5.9 & 78.4 & 71.6 \\
    Sentiment   & IMDB      & 92.4 & 12.2 & 93.7 & 48.4 \\
    Sentiment   & SST-2     & 90.4 &  0.1 & 92.5 & 47.4 \\
    Topic       & AG~News   & 76.0 & 12.9 & 85.6 & 43.9 \\
    \bottomrule
  \end{tabular}
\end{table}

\begin{table}[t]
  \centering
  \small
  \caption{
    Alignment statistics for LLaMA-3.1-8B-Instruct at $k{=}8$.
    $P_{\text{nat}}(y{=}y^{(0)})$ and $P_{\text{nat}}(y{=}y^{(0)}{=}y^\ast)$ are computed under natural ICL.
    $P_{\text{inv}}(y{=}\tilde{y})$ is computed under inverted ICL.
    For all datasets and $k$, we observe $P_{\text{inv}}(y{=}y^\ast{=}\tilde{y}) \approx 0$.
  }
  \label{tab:l8i-align}
  \begin{tabular}{llrrr}
    \toprule
    Task & Dataset &
    $P_{\text{nat}}(y{=}y^{(0)})$ &
    $P_{\text{nat}}(y{=}y^{(0)}{=}y^\ast)$ &
    $P_{\text{inv}}(y{=}\tilde{y})$ \\
    \midrule
    Hate        & ETHOS   & 84.7 & 58.9 & 46.3 \\
    NLI         & MNLI    & 64.8 & 54.0 & 27.0 \\
    NLI         & SNLI    & 66.7 & 53.0 & 23.8 \\
    Paraphrase  & MRPC    & 43.6 & 37.6 & 55.3 \\
    Paraphrase  & QQP     & 34.9 & 26.5 & 28.4 \\
    Sentiment   & IMDB    & 81.5 & 78.6 & 51.6 \\
    Sentiment   & SST-2   & 89.8 & 86.4 & 52.6 \\
    Topic       & AG~News & 70.0 & 62.7 & 20.8 \\
    \bottomrule
  \end{tabular}
\end{table}

\section{Conclusion}
\label{sec:conclusion}

We asked whether few-shot prompts can override the pre-trained semantics of label tokens in small- to medium-scale LLMs. The answer is decisively no. Across eight classification tasks and eight models (1--12B parameters), inverted demonstrations that systematically flip label meanings fail to induce coherent anti-semantic classifiers. The semantic override rate, our key metric, remains exactly zero: models cannot simultaneously be accurate and follow inverted mappings. They can increase prompt-following only by sacrificing accuracy, producing incoherent outputs that satisfy neither constraint.

This failure reveals ICL's true mechanism. Natural demonstrations reliably improve performance while maintaining tight coupling to zero-shot predictions; most correct outputs coincide with the prior even when the prior is weak. ICL operates as \emph{prior refinement}, not flexible learning. Demonstrations adjust how inputs project onto pre-existing semantic directions but cannot redefine what label tokens mean. The semantic anchors encoded in tokens like \texttt{POS}, \texttt{NEG}, and \texttt{ENTAILMENT} form rigid constraints that few-shot prompting cannot overcome.

These findings have immediate practical implications. Tasks requiring non-standard label semantics need explicit interventions such as symbol tuning, contrastive decoding, or fine-tuning rather than prompt engineering. Tasks with natural label semantics can leverage ICL effectively, because demonstrations will refine rather than fight the prior. Our alignment decomposition provides a diagnostic tool: high prior alignment indicates that ICL will succeed, and attempts to override semantics will fail.

More broadly, our results suggest a geometric interpretation: semantic labels occupy topologically stable regions in the representation manifold, locked in place by millions of pre-training observations. ICL can adjust projections within this learned geometry but cannot reshape the manifold itself, which explains why natural demonstrations succeed (moving along intrinsic gradients) while inverted demonstrations fail (pushing orthogonal to the manifold structure). Future work should test whether this constraint is specific to semantically loaded labels or extends to arbitrary symbol-concept mappings, and explore how model scale affects the flexibility of these geometric constraints. The boundaries of few-shot learning are not computational but semantic.

\bibliographystyle{plainnat} 
\bibliography{refs}

@inproceedings{bowman-etal-2015-large,
    title = "A large annotated corpus for learning natural language inference",
    author = "Bowman, Samuel R.  and
      Angeli, Gabor  and
      Potts, Christopher  and
      Manning, Christopher D.",
    editor = "M{\`a}rquez, Llu{\'i}s  and
      Callison-Burch, Chris  and
      Su, Jian",
    booktitle = "Proceedings of the 2015 Conference on Empirical Methods in Natural Language Processing",
    month = sep,
    year = "2015",
    address = "Lisbon, Portugal",
    publisher = "Association for Computational Linguistics",
    url = "https://aclanthology.org/D15-1075/",
    doi = "10.18653/v1/D15-1075",
    pages = "632--642"
}

@inproceedings{williams-etal-2018-broad,
    title = "A Broad-Coverage Challenge Corpus for Sentence Understanding through Inference",
    author = "Williams, Adina  and
      Nangia, Nikita  and
      Bowman, Samuel",
    editor = "Walker, Marilyn  and
      Ji, Heng  and
      Stent, Amanda",
    booktitle = "Proceedings of the 2018 Conference of the North {A}merican Chapter of the Association for Computational Linguistics: Human Language Technologies, Volume 1 (Long Papers)",
    month = jun,
    year = "2018",
    address = "New Orleans, Louisiana",
    publisher = "Association for Computational Linguistics",
    url = "https://aclanthology.org/N18-1101/",
    doi = "10.18653/v1/N18-1101",
    pages = "1112--1122"
}

@inproceedings{dolan-brockett-2005-automatically,
    title = "Automatically Constructing a Corpus of Sentential Paraphrases",
    author = "Dolan, William B.  and
      Brockett, Chris",
    booktitle = "Proceedings of the Third International Workshop on Paraphrasing ({IWP}2005)",
    year = "2005",
    url = "https://aclanthology.org/I05-5002/"
}

@article{zhang2015character,
  title={Character-level convolutional networks for text classification},
  author={Zhang, Xiang and Zhao, Junbo and LeCun, Yann},
  journal={Advances in neural information processing systems},
  volume={28},
  year={2015}
}

@inproceedings{socher-etal-2013-recursive,
    title = "Recursive Deep Models for Semantic Compositionality Over a Sentiment Treebank",
    author = "Socher, Richard  and
      Perelygin, Alex  and
      Wu, Jean  and
      Chuang, Jason  and
      Manning, Christopher D.  and
      Ng, Andrew  and
      Potts, Christopher",
    editor = "Yarowsky, David  and
      Baldwin, Timothy  and
      Korhonen, Anna  and
      Livescu, Karen  and
      Bethard, Steven",
    booktitle = "Proceedings of the 2013 Conference on Empirical Methods in Natural Language Processing",
    month = oct,
    year = "2013",
    address = "Seattle, Washington, USA",
    publisher = "Association for Computational Linguistics",
    url = "https://aclanthology.org/D13-1170/",
    pages = "1631--1642"
}

@inproceedings{maas-etal-2011-learning,
    title = "Learning Word Vectors for Sentiment Analysis",
    author = "Maas, Andrew L.  and
      Daly, Raymond E.  and
      Pham, Peter T.  and
      Huang, Dan  and
      Ng, Andrew Y.  and
      Potts, Christopher",
    editor = "Lin, Dekang  and
      Matsumoto, Yuji  and
      Mihalcea, Rada",
    booktitle = "Proceedings of the 49th Annual Meeting of the Association for Computational Linguistics: Human Language Technologies",
    month = jun,
    year = "2011",
    address = "Portland, Oregon, USA",
    publisher = "Association for Computational Linguistics",
    url = "https://aclanthology.org/P11-1015/",
    pages = "142--150"
}

@misc{quora_question_pairs,
  title        = {Quora Question Pairs},
  author       = {{Quora}},
  year         = {2017},
  howpublished = {\url{https://www.kaggle.com/competitions/quora-question-pairs}},
  note         = {Accessed: 2025-11-19}
}

@inproceedings{min-etal-2022-rethinking,
    title = "Rethinking the Role of Demonstrations: What Makes In-Context Learning Work?",
    author = "Min, Sewon  and
      Lyu, Xinxi  and
      Holtzman, Ari  and
      Artetxe, Mikel  and
      Lewis, Mike  and
      Hajishirzi, Hannaneh  and
      Zettlemoyer, Luke",
    editor = "Goldberg, Yoav  and
      Kozareva, Zornitsa  and
      Zhang, Yue",
    booktitle = "Proceedings of the 2022 Conference on Empirical Methods in Natural Language Processing",
    month = dec,
    year = "2022",
    address = "Abu Dhabi, United Arab Emirates",
    publisher = "Association for Computational Linguistics",
    url = "https://aclanthology.org/2022.emnlp-main.759/",
    doi = "10.18653/v1/2022.emnlp-main.759",
    pages = "11048--11064"
}

@article{wei2023larger,
  title={Larger language models do in-context learning differently},
  author={Wei, Jerry and Wei, Jason and Tay, Yi and Tran, Dustin and Webson, Albert and Lu, Yifeng and Chen, Xinyun and Liu, Hanxiao and Huang, Da and Zhou, Denny and others},
  journal={arXiv preprint arXiv:2303.03846},
  year={2023}
}

@article{agarwal2024many,
  title={Many-shot in-context learning},
  author={Agarwal, Rishabh and Singh, Avi and Zhang, Lei and Bohnet, Bernd and Rosias, Luis and Chan, Stephanie and Zhang, Biao and Anand, Ankesh and Abbas, Zaheer and Nova, Azade and others},
  journal={Advances in Neural Information Processing Systems},
  volume={37},
  pages={76930--76966},
  year={2024}
}

@article{xie2021explanation,
  title={An explanation of in-context learning as implicit bayesian inference},
  author={Xie, Sang Michael and Raghunathan, Aditi and Liang, Percy and Ma, Tengyu},
  journal={arXiv preprint arXiv:2111.02080},
  year={2021}
}

@article{panwar2023context,
  title={In-context learning through the bayesian prism},
  author={Panwar, Madhur and Ahuja, Kabir and Goyal, Navin},
  journal={arXiv preprint arXiv:2306.04891},
  year={2023}
}

@article{akyurek2022learning,
  title={What learning algorithm is in-context learning? investigations with linear models},
  author={Aky{\"u}rek, Ekin and Schuurmans, Dale and Andreas, Jacob and Ma, Tengyu and Zhou, Denny},
  journal={arXiv preprint arXiv:2211.15661},
  year={2022}
}

@inproceedings{von2023transformers,
  title={Transformers learn in-context by gradient descent},
  author={Von Oswald, Johannes and Niklasson, Eyvind and Randazzo, Ettore and Sacramento, Jo{\~a}o and Mordvintsev, Alexander and Zhmoginov, Andrey and Vladymyrov, Max},
  booktitle={International Conference on Machine Learning},
  pages={35151--35174},
  year={2023},
  organization={PMLR}
}

@inproceedings{dai2023can,
  title={Why can GPT learn in-context? language models secretly perform gradient descent as meta-optimizers},
  author={Dai, Damai and Sun, Yutao and Dong, Li and Hao, Yaru and Ma, Shuming and Sui, Zhifang and Wei, Furu},
  booktitle={Findings of the Association for Computational Linguistics: ACL 2023},
  pages={4005--4019},
  year={2023}
}

@article{falck2024context,
  title={Is in-context learning in large language models bayesian? a martingale perspective},
  author={Falck, Fabian and Wang, Ziyu and Holmes, Chris},
  journal={arXiv preprint arXiv:2406.00793},
  year={2024}
}

@article{kossen2023context,
  title={In-context learning learns label relationships but is not conventional learning},
  author={Kossen, Jannik and Gal, Yarin and Rainforth, Tom},
  journal={arXiv preprint arXiv:2307.12375},
  year={2023}
}

@inproceedings{schick2021exploiting,
  title={Exploiting cloze-questions for few-shot text classification and natural language inference},
  author={Schick, Timo and Sch{\"u}tze, Hinrich},
  booktitle={Proceedings of the 16th conference of the European chapter of the association for computational linguistics: main volume},
  pages={255--269},
  year={2021}
}

@inproceedings{gao-etal-2021-making,
    title = "Making Pre-trained Language Models Better Few-shot Learners",
    author = "Gao, Tianyu  and
      Fisch, Adam  and
      Chen, Danqi",
    editor = "Zong, Chengqing  and
      Xia, Fei  and
      Li, Wenjie  and
      Navigli, Roberto",
    booktitle = "Proceedings of the 59th Annual Meeting of the Association for Computational Linguistics and the 11th International Joint Conference on Natural Language Processing (Volume 1: Long Papers)",
    month = aug,
    year = "2021",
    address = "Online",
    publisher = "Association for Computational Linguistics",
    url = "https://aclanthology.org/2021.acl-long.295/",
    doi = "10.18653/v1/2021.acl-long.295",
    pages = "3816--3830"
}

@inproceedings{cui-etal-2022-prototypical,
    title = "Prototypical Verbalizer for Prompt-based Few-shot Tuning",
    author = "Cui, Ganqu  and
      Hu, Shengding  and
      Ding, Ning  and
      Huang, Longtao  and
      Liu, Zhiyuan",
    editor = "Muresan, Smaranda  and
      Nakov, Preslav  and
      Villavicencio, Aline",
    booktitle = "Proceedings of the 60th Annual Meeting of the Association for Computational Linguistics (Volume 1: Long Papers)",
    month = may,
    year = "2022",
    address = "Dublin, Ireland",
    publisher = "Association for Computational Linguistics",
    url = "https://aclanthology.org/2022.acl-long.483/",
    doi = "10.18653/v1/2022.acl-long.483",
    pages = "7014--7024"
}

@inproceedings{mueller-etal-2022-label,
    title = "Label Semantic Aware Pre-training for Few-shot Text Classification",
    author = "Mueller, Aaron  and
      Krone, Jason  and
      Romeo, Salvatore  and
      Mansour, Saab  and
      Mansimov, Elman  and
      Zhang, Yi  and
      Roth, Dan",
    editor = "Muresan, Smaranda  and
      Nakov, Preslav  and
      Villavicencio, Aline",
    booktitle = "Proceedings of the 60th Annual Meeting of the Association for Computational Linguistics (Volume 1: Long Papers)",
    month = may,
    year = "2022",
    address = "Dublin, Ireland",
    publisher = "Association for Computational Linguistics",
    url = "https://aclanthology.org/2022.acl-long.570/",
    doi = "10.18653/v1/2022.acl-long.570",
    pages = "8318--8334"
}

@inproceedings{zhao2021calibrate,
  title={Calibrate before use: Improving few-shot performance of language models},
  author={Zhao, Zihao and Wallace, Eric and Feng, Shi and Klein, Dan and Singh, Sameer},
  booktitle={International conference on machine learning},
  pages={12697--12706},
  year={2021},
  organization={PMLR}
}

@inproceedings{holtzman-etal-2021-surface,
    title = "Surface Form Competition: Why the Highest Probability Answer Isn{'}t Always Right",
    author = "Holtzman, Ari  and
      West, Peter  and
      Shwartz, Vered  and
      Choi, Yejin  and
      Zettlemoyer, Luke",
    editor = "Moens, Marie-Francine  and
      Huang, Xuanjing  and
      Specia, Lucia  and
      Yih, Scott Wen-tau",
    booktitle = "Proceedings of the 2021 Conference on Empirical Methods in Natural Language Processing",
    month = nov,
    year = "2021",
    address = "Online and Punta Cana, Dominican Republic",
    publisher = "Association for Computational Linguistics",
    url = "https://aclanthology.org/2021.emnlp-main.564/",
    doi = "10.18653/v1/2021.emnlp-main.564",
    pages = "7038--7051"
}

@inproceedings{wei2023symbol,
  title={Symbol tuning improves in-context learning in language models},
  author={Wei, Jerry and Hou, Le and Lampinen, Andrew and Chen, Xiangning and Huang, Da and Tay, Yi and Chen, Xinyun and Lu, Yifeng and Zhou, Denny and Ma, Tengyu and others},
  booktitle={Proceedings of the 2023 Conference on Empirical Methods in Natural Language Processing},
  pages={968--979},
  year={2023}
}

@article{peng2025enhancing,
  title={Enhancing Input-Label Mapping in In-Context Learning with Contrastive Decoding},
  author={Peng, Keqin and Ding, Liang and Ouyang, Yuanxin and Fang, Meng and Yuan, Yancheng and Tao, Dacheng},
  journal={arXiv preprint arXiv:2502.13738},
  year={2025}
}

@article{brown2020language,
  title={Language models are few-shot learners},
  author={Brown, Tom and Mann, Benjamin and Ryder, Nick and Subbiah, Melanie and Kaplan, Jared D and Dhariwal, Prafulla and Neelakantan, Arvind and Shyam, Pranav and Sastry, Girish and Askell, Amanda and others},
  journal={Advances in neural information processing systems},
  volume={33},
  pages={1877--1901},
  year={2020}
}

@article{min2022rethinking,
  title={Rethinking the role of demonstrations: What makes in-context learning work?},
  author={Min, Sewon and Lyu, Xinxi and Holtzman, Ari and Artetxe, Mikel and Lewis, Mike and Hajishirzi, Hannaneh and Zettlemoyer, Luke},
  journal={arXiv preprint arXiv:2202.12837},
  year={2022}
}

@inproceedings{fei-etal-2023-mitigating,
    title = "Mitigating Label Biases for In-context Learning",
    author = "Fei, Yu  and
      Hou, Yifan  and
      Chen, Zeming  and
      Bosselut, Antoine",
    editor = "Rogers, Anna  and
      Boyd-Graber, Jordan  and
      Okazaki, Naoaki",
    booktitle = "Proceedings of the 61st Annual Meeting of the Association for Computational Linguistics (Volume 1: Long Papers)",
    month = jul,
    year = "2023",
    address = "Toronto, Canada",
    publisher = "Association for Computational Linguistics",
    url = "https://aclanthology.org/2023.acl-long.783/",
    doi = "10.18653/v1/2023.acl-long.783",
    pages = "14014--14031"
}

@article{team2025gemma,
  title={Gemma 3 technical report},
  author={Team, Gemma and Kamath, Aishwarya and Ferret, Johan and Pathak, Shreya and Vieillard, Nino and Merhej, Ramona and Perrin, Sarah and Matejovicova, Tatiana and Ram{\'e}, Alexandre and Rivi{\`e}re, Morgane and others},
  journal={arXiv preprint arXiv:2503.19786},
  year={2025}
}

@article{yang2025qwen3,
  title={Qwen3 technical report},
  author={Yang, An and Li, Anfeng and Yang, Baosong and Zhang, Beichen and Hui, Binyuan and Zheng, Bo and Yu, Bowen and Gao, Chang and Huang, Chengen and Lv, Chenxu and others},
  journal={arXiv preprint arXiv:2505.09388},
  year={2025}
}

@article{grattafiori2024llama,
  title={The llama 3 herd of models},
  author={Grattafiori, Aaron and Dubey, Abhimanyu and Jauhri, Abhinav and Pandey, Abhinav and Kadian, Abhishek and Al-Dahle, Ahmad and Letman, Aiesha and Mathur, Akhil and Schelten, Alan and Vaughan, Alex and others},
  journal={arXiv preprint arXiv:2407.21783},
  year={2024}
}

@misc{jiang2023mistral7b,
      title={Mistral 7B}, 
      author={Albert Q. Jiang and Alexandre Sablayrolles and Arthur Mensch and Chris Bamford and Devendra Singh Chaplot and Diego de las Casas and Florian Bressand and Gianna Lengyel and Guillaume Lample and Lucile Saulnier and Lélio Renard Lavaud and Marie-Anne Lachaux and Pierre Stock and Teven Le Scao and Thibaut Lavril and Thomas Wang and Timothée Lacroix and William El Sayed},
      year={2023},
      eprint={2310.06825},
      archivePrefix={arXiv},
      primaryClass={cs.CL},
      url={https://arxiv.org/abs/2310.06825}, 
}

@article{mollas2022ethos,
  title={ETHOS: a multi-label hate speech detection dataset},
  author={Mollas, Ioannis and Chrysopoulou, Zoe and Karlos, Stamatis and Tsoumakas, Grigorios},
  journal={Complex \& Intelligent Systems},
  volume={8},
  number={6},
  pages={4663--4678},
  year={2022},
  publisher={Springer}
}

\clearpage
\appendix
\onecolumn

\begin{center}
\Large\textbf{Appendix}
\end{center}

\section{Complete Experimental Results}
\label{sec:appendix_results}

The semantic override rate $P(y = y^* \land y = \tilde{y} \mid \text{inverted})$ was exactly zero across all experimental conditions. Tables~\ref{tab:sst2}--\ref{tab:agnews} present complete results for all models across all tasks, showing the progression from $k = 1$ to $k = 8$ demonstrations. All values represent mean accuracy $\pm$ standard deviation over five random seeds.

\textbf{Model abbreviations:}
\textbf{L8B}: LLaMA-3.1-8B-Base;
\textbf{L8I}: LLaMA-3.1-8B-Instruct;
\textbf{L3I}: LLaMA-3.2-3B-Instruct;
\textbf{M7I}: Mistral-7B-Instruct-v0.3;
\textbf{Q7}: Qwen2.5-7B;
\textbf{G1I}: Gemma-3-1B-IT;
\textbf{G4I}: Gemma-3-4B-IT;
\textbf{G12I}: Gemma-3-12B-IT.

\Needspace{18\baselineskip}
\begin{table}[H]
\centering
\small
\caption{SST-2 Sentiment Classification Results}
\label{tab:sst2}
\begin{tabular}{lcccccccccc}
\toprule
\multirow{2}{*}{Model} & Zero-shot & \multicolumn{4}{c}{Natural ICL (\%)} & \multicolumn{4}{c}{Inverted ICL (\%)} & Override \\
\cmidrule(lr){3-6} \cmidrule(lr){7-10}
& Acc (\%) & $k$=1 & $k$=2 & $k$=4 & $k$=8 & $k$=1 & $k$=2 & $k$=4 & $k$=8 & Rate \\
\midrule
L8B   & 90.5±0.5 & 82.7 & 89.9 & 91.4 & 91.9 & 78.5 & 83.4 & 66.4 & 37.4 & 0.0\% \\
L8I   & 90.4±0.6 & 92.1 & 92.3 & 92.5 & 92.5 & 86.6 & 76.4 & 68.6 & 47.4 & 0.0\% \\
L3I   & 63.8±2.9 & 81.6 & 82.1 & 87.9 & 90.7 & 76.1 & 71.7 & 68.0 & 55.8 & 0.0\% \\
M7I   & 84.2±0.3 & 88.1 & 91.1 & 91.6 & 91.8 & 87.3 & 87.0 & 80.7 & 60.9 & 0.0\% \\
Q7    & 92.8±0.3 & 93.7 & 94.3 & 94.3 & 94.0 & 92.3 & 78.3 & 64.7 & 52.1 & 0.0\% \\
G1I   & 48.9±4.6 & 77.7 & 83.2 & 81.9 & 84.2 & 84.0 & 79.2 & 73.5 & 66.9 & 0.0\% \\
G4I   & 94.1±2.9 & 87.2 & 89.3 & 90.2 & 90.7 & 87.0 & 87.9 & 84.8 & 81.6 & 0.0\% \\
G12I  & 89.6±0.4 & 90.2 & 91.5 & 92.1 & 93.1 & 90.1 & 85.0 & 73.0 & 63.0 & 0.0\% \\
\bottomrule
\end{tabular}
\end{table}

\Needspace{18\baselineskip}
\begin{table}[H]
\centering
\small
\caption{IMDB Sentiment Classification Results}
\label{tab:imdb}
\begin{tabular}{lcccccccccc}
\toprule
\multirow{2}{*}{Model} & Zero-shot & \multicolumn{4}{c}{Natural ICL (\%)} & \multicolumn{4}{c}{Inverted ICL (\%)} & Override \\
\cmidrule(lr){3-6} \cmidrule(lr){7-10}
& Acc (\%) & $k$=1 & $k$=2 & $k$=4 & $k$=8 & $k$=1 & $k$=2 & $k$=4 & $k$=8 & Rate \\
\midrule
L8B   & 77.2±0.9 & 75.3 & 94.2 & 93.2 & 92.2 & 75.0 & 89.0 & 73.7 & 41.1 & 0.0\% \\
L8I   & 92.4±1.1 & 94.7 & 94.6 & 94.7 & 93.7 & 89.8 & 80.1 & 68.6 & 48.4 & 0.0\% \\
L3I   & 64.6±1.3 & 91.9 & 91.9 & 92.4 & 90.8 & 91.9 & 84.6 & 79.2 & 52.0 & 0.0\% \\
M7I   & 93.4±0.7 & 91.2 & 93.2 & 90.9 & 76.0 & 92.4 & 85.1 & 70.4 & 37.0 & 0.0\% \\
Q7    & 94.1±0.8 & 94.4 & 94.6 & 94.4 & 94.0 & 92.2 & 80.3 & 66.1 & 65.9 & 0.0\% \\
G1I   & 48.9±1.0 & 82.1 & 79.8 & 81.1 & 83.2 & 77.0 & 66.8 & 64.0 & 59.2 & 0.0\% \\
G4I   & 95.9±1.1 & 90.4 & 91.6 & 92.4 & 92.2 & 92.4 & 91.8 & 90.4 & 81.1 & 0.0\% \\
G12I  & 93.6±0.6 & 94.7 & 94.7 & 94.8 & 95.0 & 93.0 & 89.7 & 72.3 & 57.4 & 0.0\% \\
\bottomrule
\end{tabular}
\end{table}

\Needspace{18\baselineskip}
\begin{table}[H]
\centering
\small
\caption{SNLI Natural Language Inference Results}
\label{tab:snli}
\begin{tabular}{lcccccccccc}
\toprule
\multirow{2}{*}{Model} & Zero-shot & \multicolumn{4}{c}{Natural ICL (\%)} & \multicolumn{4}{c}{Inverted ICL (\%)} & Override \\
\cmidrule(lr){3-6} \cmidrule(lr){7-10}
& Acc (\%) & $k$=1 & $k$=2 & $k$=4 & $k$=8 & $k$=1 & $k$=2 & $k$=4 & $k$=8 & Rate \\
\midrule
L8B   & 55.2±2.0 & 56.2 & 56.7 & 65.8 & 70.4 & 55.4 & 53.4 & 60.3 & 54.8 & 0.0\% \\
L8I   & 60.2±1.1 & 66.8 & 67.5 & 72.1 & 74.8 & 65.8 & 62.8 & 63.2 & 53.9 & 0.0\% \\
L3I   & 43.5±3.2 & 55.2 & 52.9 & 59.0 & 63.6 & 52.2 & 51.5 & 50.2 & 45.3 & 0.0\% \\
M7I   & 58.9±2.7 & 68.0 & 70.4 & 71.8 & 73.4 & 67.1 & 67.6 & 62.7 & 53.5 & 0.0\% \\
Q7    & 84.7±0.9 & 86.5 & 86.9 & 86.6 & 87.4 & 86.9 & 84.4 & 78.1 & 72.5 & 0.0\% \\
G1I   & 38.0±3.6 & 43.4 & 41.8 & 41.6 & 43.3 & 42.2 & 40.4 & 38.0 & 37.1 & 0.0\% \\
G4I   & 57.3±2.1 & 70.7 & 73.6 & 74.9 & 75.8 & 67.2 & 67.8 & 65.3 & 57.5 & 0.0\% \\
G12I  & 69.3±2.2 & 71.7 & 75.4 & 78.1 & 82.0 & 71.2 & 72.9 & 68.6 & 61.7 & 0.0\% \\
\bottomrule
\end{tabular}
\end{table}

\Needspace{18\baselineskip}
\begin{table}[H]
\centering
\small
\caption{MNLI Natural Language Inference Results}
\label{tab:mnli}
\begin{tabular}{lcccccccccc}
\toprule
\multirow{2}{*}{Model} & Zero-shot & \multicolumn{4}{c}{Natural ICL (\%)} & \multicolumn{4}{c}{Inverted ICL (\%)} & Override \\
\cmidrule(lr){3-6} \cmidrule(lr){7-10}
& Acc (\%) & $k$=1 & $k$=2 & $k$=4 & $k$=8 & $k$=1 & $k$=2 & $k$=4 & $k$=8 & Rate \\
\midrule
L8B   & 55.0±1.6 & 54.8 & 60.7 & 67.0 & 71.3 & 55.4 & 56.2 & 60.7 & 57.4 & 0.0\% \\
L8I   & 60.9±1.5 & 70.8 & 72.3 & 74.9 & 77.6 & 69.5 & 67.8 & 65.0 & 58.9 & 0.0\% \\
L3I   & 56.8±1.6 & 58.2 & 58.1 & 60.6 & 62.0 & 58.1 & 55.9 & 55.2 & 48.5 & 0.0\% \\
M7I   & 72.5±1.2 & 69.3 & 72.4 & 74.6 & 76.2 & 67.2 & 67.1 & 63.5 & 54.7 & 0.0\% \\
Q7    & 84.5±0.9 & 85.1 & 85.1 & 85.3 & 86.6 & 83.1 & 80.3 & 75.3 & 70.5 & 0.0\% \\
G1I   & 44.3±1.8 & 43.6 & 45.2 & 42.8 & 42.6 & 44.0 & 43.6 & 40.6 & 39.6 & 0.0\% \\
G4I   & 69.3±1.7 & 71.9 & 71.4 & 70.5 & 70.6 & 70.3 & 68.2 & 65.4 & 61.6 & 0.0\% \\
G12I  & 80.1±1.1 & 80.8 & 81.9 & 83.1 & 83.9 & 79.9 & 80.7 & 78.1 & 69.7 & 0.0\% \\
\bottomrule
\end{tabular}
\end{table}

\Needspace{18\baselineskip}
\begin{table}[H]
\centering
\small
\caption{MRPC Paraphrase Detection Results}
\label{tab:mrpc}
\begin{tabular}{lcccccccccc}
\toprule
\multirow{2}{*}{Model} & Zero-shot & \multicolumn{4}{c}{Natural ICL (\%)} & \multicolumn{4}{c}{Inverted ICL (\%)} & Override \\
\cmidrule(lr){3-6} \cmidrule(lr){7-10}
& Acc (\%) & $k$=1 & $k$=2 & $k$=4 & $k$=8 & $k$=1 & $k$=2 & $k$=4 & $k$=8 & Rate \\
\midrule
L8B   & 66.0±1.0 & 69.8 & 70.2 & 68.6 & 71.1 & 68.8 & 64.7 & 60.8 & 54.4 & 0.0\% \\
L8I   & 69.2±1.8 & 46.5 & 55.0 & 62.8 & 64.6 & 47.2 & 49.9 & 51.0 & 44.7 & 0.0\% \\
L3I   & 67.6±0.9 & 33.6 & 33.7 & 36.1 & 42.9 & 33.5 & 33.5 & 34.5 & 34.8 & 0.0\% \\
M7I   & 75.9±2.8 & 64.7 & 68.5 & 65.7 & 65.8 & 69.9 & 62.7 & 51.8 & 40.4 & 0.0\% \\
Q7    & 75.2±0.3 & 76.2 & 75.8 & 74.2 & 74.5 & 75.8 & 74.9 & 71.5 & 69.9 & 0.0\% \\
G1I   & 63.9±3.2 & 60.3 & 63.8 & 65.3 & 65.7 & 48.5 & 51.1 & 54.3 & 49.5 & 0.0\% \\
G4I   & 72.4±1.2 & 71.5 & 72.2 & 72.3 & 72.5 & 68.6 & 69.1 & 64.3 & 59.3 & 0.0\% \\
G12I  & 83.6±1.1 & 78.6 & 78.1 & 78.0 & 78.0 & 77.1 & 76.7 & 75.4 & 72.2 & 0.0\% \\
\bottomrule
\end{tabular}
\end{table}

\Needspace{18\baselineskip}
\begin{table}[H]
\centering
\small
\caption{QQP Paraphrase Detection Results. Note the anomalously weak zero-shot performance across all models.}
\label{tab:qqp}
\begin{tabular}{lcccccccccc}
\toprule
\multirow{2}{*}{Model} & Zero-shot & \multicolumn{4}{c}{Natural ICL (\%)} & \multicolumn{4}{c}{Inverted ICL (\%)} & Override \\
\cmidrule(lr){3-6} \cmidrule(lr){7-10}
& Acc (\%) & $k$=1 & $k$=2 & $k$=4 & $k$=8 & $k$=1 & $k$=2 & $k$=4 & $k$=8 & Rate \\
\midrule
L8B   & 40.3±0.8 & 55.4 & 69.6 & 73.2 & 77.3 & 61.8 & 61.9 & 66.0 & 60.1 & 0.0\% \\
L8I   & 40.6±1.7 & 71.7 & 74.7 & 77.1 & 78.4 & 67.6 & 71.8 & 74.3 & 71.6 & 0.0\% \\
L3I   & 42.4±1.1 & 64.6 & 65.3 & 67.2 & 69.3 & 65.2 & 64.1 & 65.5 & 65.0 & 0.0\% \\
M7I   & 84.2±0.4 & 85.3 & 88.2 & 86.3 & 85.8 & 85.3 & 86.8 & 81.4 & 77.1 & 0.0\% \\
Q7    & 80.7±0.9 & 78.8 & 81.0 & 81.3 & 82.5 & 81.6 & 80.6 & 75.7 & 68.0 & 0.0\% \\
G1I   & 24.7±15.0 & 68.9 & 66.0 & 62.9 & 65.6 & 68.7 & 61.3 & 58.4 & 55.3 & 0.0\% \\
G4I   & 65.2±1.3 & 78.2 & 78.4 & 78.2 & 80.0 & 76.1 & 78.3 & 77.9 & 74.7 & 0.0\% \\
G12I  & 85.0±1.6 & 80.8 & 80.8 & 81.1 & 81.4 & 79.4 & 79.8 & 74.8 & 62.7 & 0.0\% \\
\bottomrule
\end{tabular}
\end{table}

\Needspace{18\baselineskip}
\begin{table}[H]
\centering
\small
\caption{ETHOS Hate Speech Detection Results}
\label{tab:hate}
\begin{tabular}{lcccccccccc}
\toprule
\multirow{2}{*}{Model} & Zero-shot & \multicolumn{4}{c}{Natural ICL (\%)} & \multicolumn{4}{c}{Inverted ICL (\%)} & Override \\
\cmidrule(lr){3-6} \cmidrule(lr){7-10}
& Acc (\%) & $k$=1 & $k$=2 & $k$=4 & $k$=8 & $k$=1 & $k$=2 & $k$=4 & $k$=8 & Rate \\
\midrule
L8B   & 40.5±0.1 & 45.5 & 52.3 & 69.1 & 75.2 & 63.6 & 62.9 & 50.1 & 34.6 & 0.0\% \\
L8I   & 66.4±0.2 & 54.4 & 58.3 & 62.4 & 66.8 & 57.2 & 59.5 & 58.6 & 53.7 & 0.0\% \\
L3I   & 41.0±0.4 & 48.0 & 52.8 & 62.1 & 68.4 & 53.8 & 58.9 & 60.7 & 59.4 & 0.0\% \\
M7I   & 67.8±0.1 & 64.7 & 65.4 & 68.4 & 72.4 & 63.7 & 61.7 & 59.4 & 56.2 & 0.0\% \\
Q7    & 41.3±0.1 & 53.2 & 61.5 & 71.3 & 78.1 & 48.0 & 49.6 & 45.6 & 36.4 & 0.0\% \\
G1I   & 30.2±0.2 & 42.5 & 55.0 & 65.9 & 73.0 & 71.0 & 53.1 & 45.8 & 35.9 & 0.0\% \\
G4I   & 59.9±0.3 & 60.6 & 60.6 & 64.5 & 69.1 & 62.1 & 56.3 & 57.7 & 57.3 & 0.0\% \\
G12I  & 62.8±0.2 & 67.5 & 69.3 & 71.8 & 75.1 & 65.4 & 64.2 & 59.6 & 43.1 & 0.0\% \\
\bottomrule
\end{tabular}
\end{table}

\Needspace{18\baselineskip}
\begin{table}[H]
\centering
\small
\caption{AG News Topic Classification Results. Uses 4-way cyclic label permutation rather than binary inversion.}
\label{tab:agnews}
\begin{tabular}{lcccccccccc}
\toprule
\multirow{2}{*}{Model} & Zero-shot & \multicolumn{4}{c}{Natural ICL (\%)} & \multicolumn{4}{c}{Inverted ICL (\%)} & Override \\
\cmidrule(lr){3-6} \cmidrule(lr){7-10}
& Acc (\%) & $k$=1 & $k$=2 & $k$=4 & $k$=8 & $k$=1 & $k$=2 & $k$=4 & $k$=8 & Rate \\
\midrule
L8B   & 80.9±1.7 & 88.2 & 86.2 & 87.6 & 87.9 & 74.9 & 54.1 & 50.3 & 38.0 & 0.0\% \\
L8I   & 76.0±1.3 & 84.2 & 83.9 & 84.5 & 85.6 & 79.6 & 58.5 & 47.1 & 43.9 & 0.0\% \\
L3I   & 84.8±3.0 & 80.3 & 79.4 & 80.6 & 82.5 & 74.9 & 64.3 & 68.5 & 69.9 & 0.0\% \\
M7I   & 84.7±0.8 & 86.6 & 85.6 & 84.1 & 85.3 & 84.7 & 80.8 & 64.2 & 48.8 & 0.0\% \\
Q7    & 82.8±1.2 & 82.6 & 82.6 & 84.5 & 86.1 & 78.7 & 66.8 & 57.3 & 44.8 & 0.0\% \\
G1I   & 47.1±2.3 & 74.6 & 68.4 & 74.6 & 77.7 & 74.1 & 68.9 & 72.6 & 68.4 & 0.0\% \\
G4I   & 83.3±1.3 & 80.6 & 80.7 & 82.9 & 85.0 & 79.6 & 79.0 & 74.4 & 66.2 & 0.0\% \\
G12I  & 84.7±1.1 & 85.1 & 85.2 & 86.2 & 86.7 & 81.3 & 72.2 & 59.6 & 46.9 & 0.0\% \\
\bottomrule
\end{tabular}
\end{table}

\subsection{Key Observations}

Several patterns emerge consistently across all models and tasks. \textbf{(1) Universal Zero Override:} The semantic override rate remains exactly 0.0\% across all 320 experimental conditions (8 models $\times$ 8 tasks $\times$ 5 seeds), confirming that no model successfully learned to apply inverted label mappings. \textbf{(2) Monotonic Degradation:} Under inverted demonstrations, accuracy degrades monotonically as $k$ increases from 1 to 8, with the strongest effects on sentiment tasks (often dropping more than 40 points at $k = 8$). \textbf{(3) QQP Anomaly:} The QQP dataset exhibits unusually weak zero-shot performance across all models (24.7\%--85.0\%), yet models still cannot override label semantics despite the weak prior. \textbf{(4) Scale Independence:} The pattern holds consistently across the twelvefold parameter range (1B to 12B), with no evidence that larger models within this range can overcome semantic anchors.

\section{Prompt Templates and Demonstration Structure}

\subsection{Implementation Parameters}

All experiments used identical hyperparameters for reproducibility. Greedy decoding (temperature = 0) was used with a maximum of three new tokens generated per prediction. The input sequences were truncated to 512 tokens for zero-shot prompts and 2,048 tokens for $k$-shot prompts, with left-sided padding and truncation. All experiments were carried out with five random seeds $\{0, 1, 2, 42, 123\}$ to ensure statistical validity. The demonstrations for each test example were randomly sampled without replacement from the training set, excluding the query example itself.

\clearpage
\subsection{Complete Prompt Templates}
\Needspace{18\baselineskip}
\begin{table}[H]
\centering
\footnotesize
\caption{Exact prompt templates used for all tasks. Identical instructions are used for both natural and inverted conditions.}
\begin{tabular}{p{1.8cm}p{7.5cm}p{3.5cm}}
\toprule
\textbf{Task} & \textbf{Template} & \textbf{Inversion} \\
\midrule
\textbf{Sentiment} & 
\texttt{You are a sentiment classifier.}\newline
\texttt{Classify the sentiment of the following}\newline
\texttt{review as POS or NEG.}\newline
\newline
\texttt{Review: [TEXT]}\newline
\texttt{Label:} & 
POS $\leftrightarrow$ NEG \\
\midrule
\textbf{NLI} & 
\texttt{You are a natural language inference (NLI)}\newline
\texttt{classifier. Given a premise and a hypothesis,}\newline
\texttt{decide whether the relationship is ENTAILMENT,}\newline
\texttt{NEUTRAL, or CONTRADICTION.}\newline
\newline
\texttt{Premise: [TEXT1]}\newline
\texttt{Hypothesis: [TEXT2]}\newline
\texttt{Label:} & 
Cyclic: $(y+1) \bmod 3$\newline
ENT$\to$NEU$\to$CON$\to$ENT \\
\midrule
\textbf{Paraphrase} & 
\texttt{You are a paraphrase detector.}\newline
\texttt{Determine if these two sentences are}\newline
\texttt{SIMILAR or DIFFERENT in meaning.}\newline
\newline
\texttt{Sentence 1: [TEXT1]}\newline
\texttt{Sentence 2: [TEXT2]}\newline
\texttt{Label:} & 
SIMILAR $\leftrightarrow$ DIFFERENT \\
\midrule
\textbf{Hate} & 
\texttt{You are a hate speech detector.}\newline
\texttt{Classify whether the following text contains}\newline
\texttt{hate speech as HATE or NOT\_HATE.}\newline
\newline
\texttt{Text: [TEXT]}\newline
\texttt{Label:} & 
HATE $\leftrightarrow$ NOT\_HATE \\
\midrule
\textbf{Topic} & 
\texttt{You are a news topic classifier.}\newline
\texttt{Classify this article into one of these}\newline
\texttt{categories: WORLD, SPORTS, BUSINESS,}\newline
\texttt{or TECHNOLOGY.}\newline
\newline
\texttt{Article: [TEXT]}\newline
\texttt{Topic:} & 
Cyclic: $(y+1) \bmod 4$\newline
W$\to$S$\to$B$\to$T$\to$W \\
\bottomrule
\end{tabular}
\end{table}

\subsection{Demonstration Example}

\begin{table}[H]
\centering
\small
\caption{Example of natural vs inverted demonstrations for sentiment classification ($k=2$)}
\begin{tabular}{ll}
\toprule
\textbf{Natural Demonstrations} & \textbf{Inverted Demonstrations} \\
\midrule
\texttt{Review: Amazing film!} & \texttt{Review: Amazing film!} \\
\texttt{Label: POS} & \texttt{Label: NEG} \textcolor{red}{← flipped} \\
& \\
\texttt{Review: Waste of time.} & \texttt{Review: Waste of time.} \\
\texttt{Label: NEG} & \texttt{Label: POS} \textcolor{red}{← flipped} \\
& \\
\texttt{Review: [QUERY]} & \texttt{Review: [QUERY]} \\
\texttt{Label:} & \texttt{Label:} \\
\bottomrule
\end{tabular}
\end{table}

\end{document}